\documentclass[conference]{IEEEtran}
\IEEEoverridecommandlockouts
\usepackage{cite}
\usepackage{amsmath,amssymb,amsfonts}
\usepackage{booktabs}
\usepackage{amsmath}
\usepackage{multirow}
\usepackage{algorithmic}
\usepackage{graphicx}
\usepackage{textcomp}
\usepackage{xcolor}
\def\BibTeX{{\rm B\kern-.05em{\sc i\kern-.025em b}\kern-.08em
    T\kern-.1667em\lower.7ex\hbox{E}\kern-.125emX}}
\begin{document}

\title{Trigger-GNN: A Trigger-Based Graph Neural Network for Nested Named Entity Recognition}
% \author{\IEEEauthorblockN{Anonymous Authors}}
\author{
    \IEEEauthorblockN{Yuan Sui$^{a}$, Fanyang Bu$^{a}$, Yingting Hu$^a$, Liang Zhang$^{b*}$, Wei Yan$^{a*}$}
    \IEEEauthorblockA{$^a$ School of Information Science and Engineering, Shandong Normal University, Jinan, China}
    \IEEEauthorblockA{$^b$ Institute of Frontier and Interdisciplinary Science and Key Laboratory of Particle Physics\\ and Particle Irradiation (MOE), Shandong University, Qingdao, China}
    % \IEEEauthorblockA{\{zhangsan\}@XXX.com, \{lisi, wangwu\}@XXX.edu.cn, {g.li}@XXX.com}
    \IEEEauthorblockA{yuansui08@gmail.com, gw76999@gmail.com, 945774816@qq.com, zhang.l@sdu.edu.cn, wyaninsa@gmail.com}
}

% \author{\IEEEauthorblockN{1\textsuperscript{st} Yuan Sui}
% \IEEEauthorblockA{\textit{Shandong Normal University} \\
% \textit{School of Information Science and Engineering}\\
% Jinan, China \\
% yuansui08@gmail.com}
% \and
% \IEEEauthorblockN{2\textsuperscript{nd} Fanyang Bu}
% \IEEEauthorblockA{\textit{Shandong Normal University} \\
% \textit{School of Information Science and Engineering}\\
% Jinan, China \\
% gw76999@gmail.com}
% \and
% \IEEEauthorblockN{3\textsuperscript{rd} Yingting Hu}
% \IEEEauthorblockA{\textit{Shandong Normal University} \\
% \textit{School of Information Science and Engineering}\\
% Jinan, China \\
% 945774816@qq.com}
% \and
% \IEEEauthorblockN{4\textsuperscript{th} Wei Yan*\footnote{* refers to the corresponding author of this paper.}}
% \IEEEauthorblockA{\textit{Shandong Normal University} \\
% \textit{School of Information Science and Engineering}\\
% Jinan, China \\
% wyaninsa@gmail.com}
% \and
% \IEEEauthorblockN{5\textsuperscript{th} Liang Zhang*}
% \IEEEauthorblockA{\textit{Shandong Normal University, Institute of Frontier and Interdisciplinary}\\
% \textit{Science and Key Laboratory of Particle Physics and Particle Irradiation (MOE)}\\
% Qingdao, China\\
% zhang.l@sdu.edu.cn}
% }

\maketitle

\begin{abstract}
Nested named entity recognition (NER) aims to identify the entity boundaries and recognize categories of the named entities in a complex hierarchical sentence. Some works have been done using character-level, word-level, or lexicon-level based models. However, such researches ignore the role of the complementary annotations. In this paper, we propose a trigger-based graph neural network (Trigger-GNN) to leverage the nested NER. 
It obtains the complementary annotation embeddings through entity trigger encoding and semantic matching, and tackle nested entity utilizing an efficient graph message passing architecture, aggregation-update mode.
We posit that using entity triggers as external annotations can add in complementary supervision signals on the whole sentences. It helps the model to learn and generalize more efficiently and cost-effectively. Experiments show that the Trigger-GNN consistently outperforms the baselines on four public NER datasets, and it can effectively alleviate the nested NER.
\end{abstract}


\section{Introduction}
Named entity recognition (nested-NER) aims to identify the entity boundaries and recognize the categories of named entities in a sentence \cite{Yadav2019,Chiu2016}. 
The categorizes belong to the pre-defined semantic types, such as person, location, organization \cite{Tran2017}.
With applications ranging from AI-based dialogue systems to combing the natural language and semantic web in learning environments, NER plays a standalone role.

\begin{figure}[ht]
	\includegraphics[width=\linewidth]{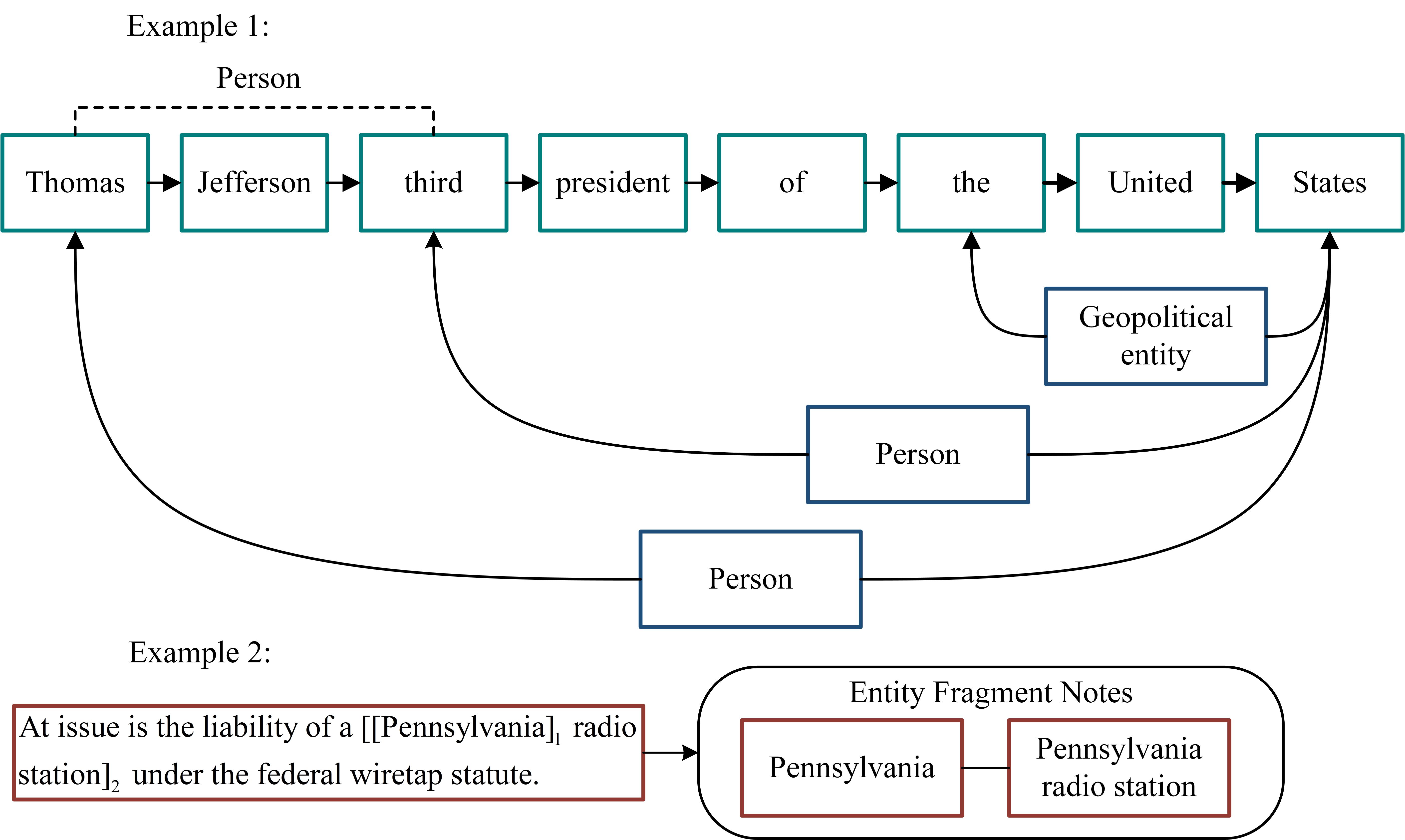}
	\caption{Examples of the nested entity recognition.}
	\label{fig:introduction}
\end{figure}

However, nested NER is always a thorny challenge due to its complex hierarchical structure.
Fig~\ref{fig:introduction} illustrates two examples of the nested entity strings. The upper example shows that ``Thomas Jefferson, third president of the United States" should be labeled jointly to constitute a complete entity statement and expressed as a person entity (PER).
However, it also involves two distinct entities: ``the United States" and ``third president of the United States", which can be expressed as a geopolitical entity (GEO) or a PER separately.
The nest problem hampers the judgment of the entities boundaries. Sometimes, it appears as an overlapping text. As shown as the bottom example of Fig~\ref{fig:introduction}, ``Pennsylvania radio station" is overlapped. 
``Pennsylvania" can be treated as a PER or concatenate with the following two words as a organization institute (ORI), ``Pennsylvania radio station''.

Generally, one intuitive way to leverage the nested NER is to stack flat NER layers \cite{Ju2018a,Ju2018,Wang2019,Nguyen2016}. Ju {\it et al.} \cite{Ju2018} proposed to recognize the nested entities by stacking the flat NER layers dynamically. It concatenates the output of each LSTM structure from the current NER layer and subsequently feeds them into the next flat NER layer. It makes full use of the information encoded in the corresponding internal entities (entities exist in the internal layers) in an inside-out manner.
However, this kind of layered model cannot reuse the outside information, which means it has a single direction for information transfer from the inner layer to the outside way. 
Hence, joint learning flat entities and their inner dependencies has attracted research attention. Luo {\it et al.} \cite{Luo2020} proposed a bipartite flat graph network considering the bidirectional delivery of information from innermost layers to outer ones. It indeed tackle the lack of the dependencies of inner entities. However, it still lack clear rationales to execute the delivery process. The nested NER is still a nontrivial task due to its complex structure.

Recent advances in nested-NER mainly focus on training an superior neural network model on different levels of the semantic hierarchy, \textit{e.g.}, character-level \cite{Huang2015}, word-level \cite{Chiu2016}, and lexicon-level \cite{Gui2019,Zhang2018}.
However, such researches ignore the role of the complementary annotation 
\footnote{Complementary annotation refers to the supplementary explanation representing the subtle reasons why humans label an entity among the sentence. For example, ``Tom \underline{traveled} a lot last year \underline{in} Silicon.'' where `Silicon' can be signed as location (LOC) entity. The reason why we make this judgment is due to the cue phrase ``travel $\ldots$ in''. Considering semantics, it reveals that there should be an LOC right behind the word `in'. Actually, this is how the complementary annotation works.}.
In recent years, some works have been done using this external supplementary explanation as compensation \cite{Lin2020}.
Lin {\it et al.} first proposed entity trigger as an complementary annotation for facilitating NER models using label-efficient learning. It defines a set of words from the sentence as the entity triggers, which brings the complementary annotation for recognizing the entities of the sentence.
For example, given a sentence ``We had a fantastic lunch at Rumble Fish yesterday, where the food is my favorite.'', where ``had $\cdots$ lunch at'' and ``where the food'' are two distinct triggers associated with the restaurant entity ``Rumble Fish''. 
These distinct triggers \footnote{To be noted that an entity trigger should follows some rules: (1) an entity trigger should involves necessary and sufficient cue for entity recognition process; (2) some modifier words should be removed, \textit{e.g.}, ``fantastic''.} explicitly indicate the location information of the candidate entities and helps to better anchor them. 
In addition, the defined entity triggers can be reused. Statements with similar phrase structure can be linked to a equivalent class and reuse the same entity trigger as well. It means the most of our commonly used sentences can be formed into some entity triggers rather than labelling all these sentences manually. It helps the model training to be cost-effectively.

Our intuition is that these triggers can add in complementary supervision on the whole sentences and thus help the nested-NER models to learn and generalize more efficiently and cost-effectively.
It utilizes the recurrent neural networks (RNN) to encode the sentences sequentially. However, the underlying structure of the sentences are not strictly serial. In fact, RNN-based models process words in a strictly sequential order, and a word have precedence when assigned to its left word. More seriously, these methods label the candidate entities only using the previous partial sequences without seeing the remain words. It lacks the capacity of capturing the long-term dependency and high-level features among the sentence.

To this end, we introduce a trigger-based graph neural network (Trigger-GNN). It casts the nested-NER into a node classification task, and breaks the serialization processing of RNNs using an recursive graph neural network. In nested NER, we have to tackle the uncertain rules for discontinuous token sequences and the confusing error from multiple rules applying to an input instance at the same time. We propose an updation-aggregation training manner to address this issue, \textit{i.e.}, the node representation is updated by aggregating the representation of its adjacent edges and the graph-level node recursively. 
The multiple iterations of aggregation enables Trigger-GNN to continuously verify the nested words based on the global context information. The key contributions of this paper can be summarized as follows:
\begin{itemize}
	\item We develop a trigger-based graph neural network for the nested NER task in a cost-effective manner, and cast the problem into a graph node classification task.
	\item We propose to capture the global context information and local compositions to tackle nested NER through a recursively aggregating mechanism.
	\item Experiments show that the Trigger-GNN is cost-effective and efficient on four public NER datasets.
\end{itemize}

\section{Related Work}

\subsection{Graph Neural Networks on Texts}
Graph neural networks have been successfully applied to several text classification tasks \cite{Li2019a,Yao2019a,Lei2021,Liu2020,Huang2019}.
Liu {\it et al.} \cite{Liu2020} proposed a tensor graph convolutional networks called Tensor-GCN. The Tensor-GCN uses the text graph tensor to describe semantic, syntactic, and sequential contextual information. The model uses the combination of the intra-graph propagation and inter-graph propagation to aggregate the information from the neighborhood nodes in a single graph, and to harmonize heterogeneous information between graphs.
Huang {\it et al.} \cite{Huang2019} proposed to build graph for each input text using sharing parameters, instead of training on a single graph for the whole corpus. It indeed removes the burden of dependence from the individual text to the entire corpus. However, it still preserves the global information.
Yao {\it et al.} \cite{Yao2019a} developed a single text graph convolutional network (Text-GCN) based on the word co-occurrence and document-word relations. It jointly learns the embeddings of both words and documents, and it is supervised by the documents' class labels.

\subsection{Nested Named Entity Recognition}
There has been a growing amount of efforts towards NER, and have been explored in several directions, including rule-based, statistics-based and deep neural network-based \cite{Yadav2019,Huang2015,Luo2020a,Wang2019,Zheng2017,Yao2019}. However, nested named entity recognition is always a thorny issue because of its complex hierarchical structure.

Towards alleviating the nested NER, recent works have mainly focus on stacking flat NER layers \cite{Ju2018a,Ju2018,Wang2019,Nguyen2016}.
Ju {\it et al.} \cite{Ju2018} proposed to recognize the nested entities by stacking flat NER layers dynamically. The model concatenates the output of each LSTM structure in the current flat NER layer to build a revolutionary representation for detected entities, and subsequently feeds them into the next flat NER layer. It makes full use of the encoded information in the corresponding internal entities in an inside-out manner. However, the layered model cannot reuse the outside information, which means it has a single direction for information transfer from the inner layer to the outside way. 

Another line of research aims to combine the bipartite graph with the flat NER layers.
Luo {\it et al.} \cite{Luo2020} proposed a bipartite flat GNN to learn the flat entities and their inner dependencies jointly. It constructs a graph module to deliver the bidirectional information from innermost graph layer to the outermost one. The leaned information carries the dependencies of inner entities and can be exploited to improve outermost entity predictions. Due to the graph module, the transfer of the information is bidirectional. The layered model can reuse the outside information as well. However, this method lacks clear and confident rationales to conduct the process of notation. While the entity triggers can bring the complementary annotation to help promote the labelling process.

Inspired by the achievements aforementioned (e.g., graph neural networks on texts~\cite{Gui2019} and learning with entity triggers~\cite{Lin2020}), we propose to use a graph neural network integrating the complementary annotation. 
The Trigger-GNN enables two significant issues: (1) multiple graph-based interactions among the words, entity triggers, and the whole sentence semantics; and (2) using trigger-based instead of lexicon-based \cite{Gui2019} can bring some complementary information as the supervision signals, and thus help the model to learn and generalize more efficiently.

\section{Trigger-Based Graph Neural Network}
\begin{figure}[ht]
	\includegraphics[width=\linewidth]{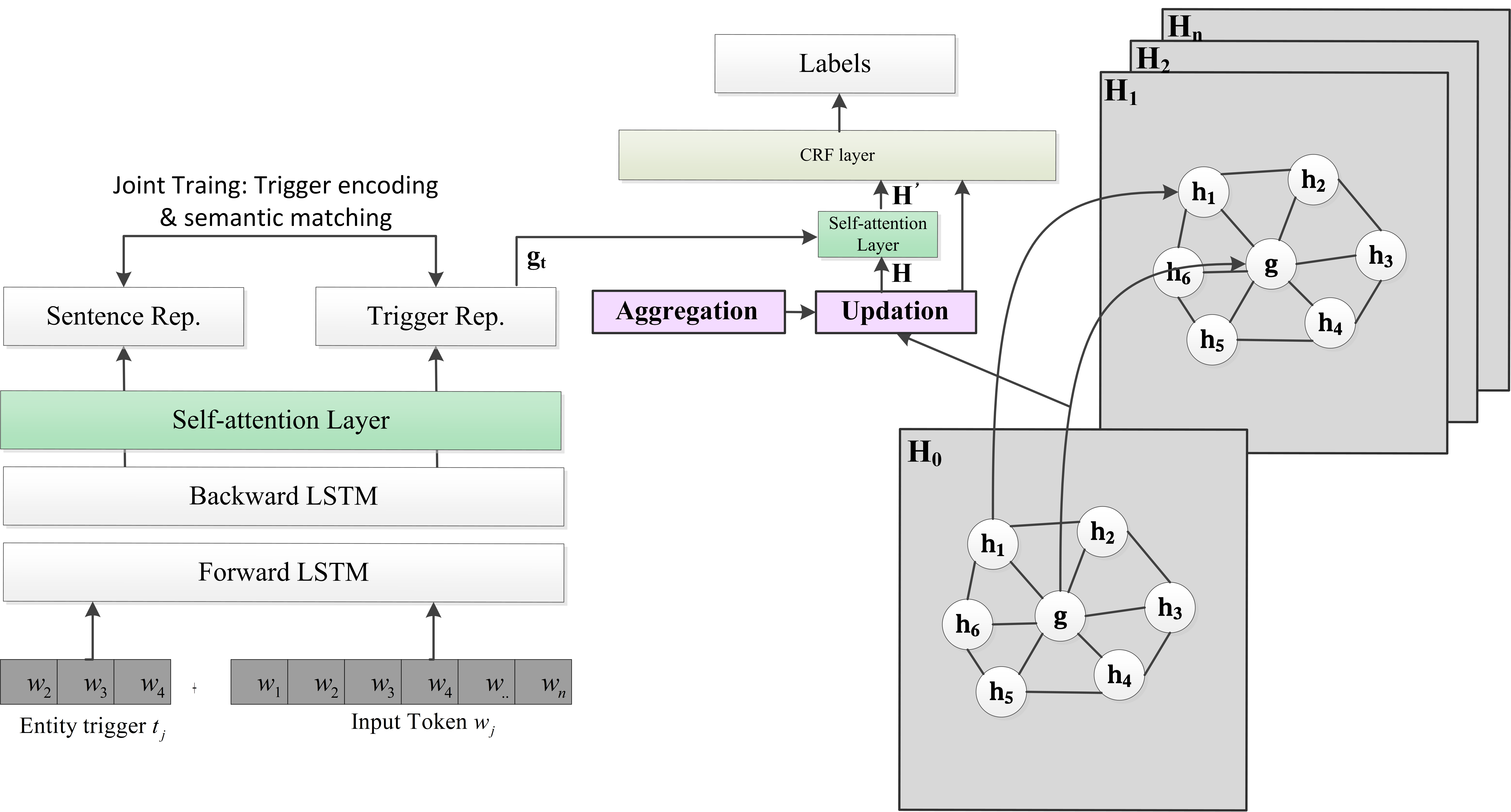}
	\caption{Trigger-based graph neural networks model. It obtains the sentence representation and trigger representation from the input entity trigger \(t_j\) and input token \(w_j\) through a bidirectional LSTM and self-attention based soft matching. We jointly train the model using both trigger encoding and semantic matching. In addition, the learned embedding will be executed through an aggregation-updation based graph neural network to better extract the interactions among the words, sentences, and the complementary annotations.}
	\label{fig:model-illustration}
\end{figure}

In this section, we detail the Trigger-GNN model. The idea of the whole model is briefly illustrated in Fig~\ref{fig:model-illustration}.
Trigger-GNN obtains the complementary annotation embeddings through entity trigger encoding and semantic matching. And it uses an efficient graph message passing architecture \cite{Li2019}, aggregation-updation mode, to better extract the interactions among the words, sentences, and complementary annotations.

\subsection{Trigger Encoding and Semantic Matching}
\label{sec:trigger-encoding and semantic-matching}
In this section, we propose to train an encoder architecture for entity trigger learning, and to match the entity trigger with its corresponding sentence using an attention-based representations. 
Our intuition is that the desired representation of the entity triggers should involve the semantics with the hidden states of the tokens from the sentence in a shared embedding space.
Specifically, for a sentence $S = w_1, w_2, \dots, w_n$ with multiple entities $e_1, e_2, \dots, e_k$, we assume that there is a set of triggers $T_i = t_1, t_2, \ldots, t_n$. 
We reformat the input of our model to enable efficient batch training. Each entity is linked to one of its corresponding trigger denoted as $(x,e_i,t_j)$, where $x$ is the tokenization list of the sentence $S$.
For each reformed training batch, we first apply a bidirectional LSTM on the sequence of word vectors of $x$ using Glove word embedding \cite{Pennington2014}. 
It returns the hidden states $h_i$ of each token $x_i$, and $\mathbf{H}$ is denoted as the matrix containing the embedding representation for all the tokens; $\mathbf{Z}$ is denoted as the matrix containing the embedding representation for all the trigger $T$.
We utilize the self-attention method introduced by \cite{Lin2020} to obtain the representation of both triggers and sentences:
\begin{equation}
	\begin{aligned}
		\mathbf{g}_{\mathrm{s}}&=\operatorname{SoftMax}\left(W_{2} \tanh \left(W_{1} \mathbf{H}^{T}\right)\right) \mathbf{H} \\
		\mathbf{g}_{\mathbf{t}} &=\operatorname{SoftMax}\left(W_{2} \tanh \left(W_{1} \mathbf{Z}^{T}\right)\right) \mathbf{Z} \\
	\end{aligned}
\end{equation}
where $W_1$ and $W_2$ are two trainable parameters in the model. $\mathbf{g}_{\mathbf{s}}$ is denoted as the final sentence's representation. It reveals the weighted sum of all the token representations in the whole sentence. Similarly, $\mathbf{g}_{\mathbf{t}}$ is denoted as the final trigger's representation.

We train the trigger encoder with the help of the type of the trigger's associated entity. We use a simple multi-class classifier to predicate the type of such entity, and the corresponding type of each entity returned is denoted as $\Omega$. The loss function of the classifier is as follows:
\begin{equation}
	\begin{aligned}
		L_1 &= -\sum log(\Omega|\mathbf{g}_{\mathbf{t}};\theta_{1})
	\end{aligned}
\end{equation}
where $\theta_{1}$ is a trainable parameter in the model.

We match the triggers and sentences by $\mathbf{g}_{\mathbf{t}}$ and $\mathbf{g}_{\mathbf{s}}$ according to Eq.\ref{eq.1}. During the training process, we mix the triggers and the sentences randomly to sample some negative samples to tackle the imbalance of the positive and negative samples. We expect a margin $m$ between the sample embeddings for the negative situation. The loss function of the semantic matching is as follows:
\begin{equation}
	\begin{aligned}
		L_2 &= \alpha\frac{1}{2}(\left||\mathbf{g}_{\mathbf{s}}-\mathbf{g}_{\mathbf{t}}\right||_{2})^2+(1-\alpha)\frac{1}{2}\{\mathtt{max}\left(0,m-\left||\mathbf{g}_{\mathbf{s}}-\mathbf{g}_{\mathbf{t}}\right||_{2}\right)\}^2
	\end{aligned}
\end{equation}
where $\alpha$ is defined to confirm whether the trigger is originally in the sentence of not. When $\alpha$ is set to $1$, the trigger is originally in the sentence; and when it is set to $0$, the trigger is not originally in the sentence.
The joint loss function of the trigger encoding and semantic matching is $L=L1+\lambda L2$, where $\lambda$ is a hyper-parameter for fine-tuning. In our experiment, we define $\lambda=1.3$.

\subsection{Text Graph Construction}
In this section, we detail how to convert the whole sentences into a directed graph. We define each word in the sentences as a node, and add the edges between each node according to the corresponding lexicon\footnote{To this end, we have to maintain a lexicon list during our experiment, find more details in the Experiment Analysis Section.} Also, we design a graph-level node to gather all the information from the nodes and edges in the text graph. The graph-level node can help the node representation by removing the ambiguity.

Formally, let $S = w_1, w_2, \dots, w_n$ denote a sentence, where $w_i$ denotes the $i$-th word. The potential lexicon $T = t_1, t_2, \dots, t_n$ matching a word sub-sequence can be formulated as $ t_{b,e} = w_b, w_{b+1}, \dots, w_{e-1}, w_e$, where the index of the first word and the last word are $b$ and $e$.
In this work, we propose to denote directed and labeled multi-graphs as $G = (V,E,U)$ with nodes (words: $v_i \in V$), labeled edges (relations: ($t_{b,e},r)\in E$, where $r\in R$ is a relation type according to the entity trigger), and graph-level node (global attributes: $u_i \in U$).
Once a word sub-sequence matches a candidate lexicon such as $t_{b,e}$, we add one edge $e_{b,e}\in E$, indicating that it starts from the beginning word $w_b$ to the ending word $w_e$. The graph-level node is to capture the global information of the entire text graph. Formally, it is represented as the sum of the representation of all the nodes and edges of the graph. For a graph with $n$ word nodes and $m$ edges, there are $m+n$ relations linking each node and edge to the shared graph-level node representation. 
In addition, we construct the transpose of the text graph $G$ according to \cite{Gui2019}. It is another directed graph denoted as $G^{\mathrm{T}}$ with the same set of nodes but all edges reversed compared to $G$. Similar to the bi-directional LSTM, we compose $G$ and $G^{\mathrm{T}}$ as a bi-directional text graph as the input and concatenate the hidden states from $G$ and $G^{\mathrm{T}}$ as the final outputs.

\subsection{Recursive Graph Neural Networks}
In this part, we detail the structure of the recursive-based graph neural network from three subsections: updation module, aggregation module, and trigger-enhanced decoding and tagging.

\subsubsection{Updation Module}
The hidden vectors of all of the tokens $\mathcal{H}$ is used as the representation for each node in the graph, where $\mathcal{H} = x_1, x_2, \ldots, x_n$. And the hidden vectors of the trigger token $\mathcal{Z}$ is used as the representation for each edge, where $\mathcal{Z} = y_1, y_2, \ldots, y_m$. 
% The recursive structure is an applied layer using gated graph neural networks \cite{Li2019}. 
Formally, the hidden state of the text graph is denoted as (at the $\ell$-th layer):
\begin{equation}\label{eq.1}
	H^{\ell} = \left<h_1^{\ell},h_2^{\ell},\ldots,h_m^{\ell},g^{\ell}\right>
\end{equation}
where $h_i$ is the hidden status of each node. $g$ represents the graph-level node.

For the initial state $H^0$ of the text graph, the hidden states of the $i$-th nodes $h_i^0$ is set to its embedding, where $h_i^0 = x_i$. The transition from $h_i^{\ell-1}$ to $h_i^{\ell}$ is calculated as follows:
\begin{equation}\label{eq.2}
	\begin{aligned}
		i_{i}^{l} &=\sigma\left(W_{i}\left[h_{i}^{l-1} ; x_{i} ; g^{l-1} ; \mathcal{N}_{i}^{l-1}\right]+b_{i}\right) \\
		f_{i}^{l} &=\sigma\left(W_{f}\left[h_{i}^{l-1} ; x_{i} ; g^{l-1} ; \mathcal{N}_{i}^{l-1}\right]+b_{f}\right) \\
		o_{i}^{l} &=\sigma\left(W_{o}\left[h_{i}^{l-1} ; x_{i} ; g^{l-1} ; \mathcal{N}_{i}^{l-1}\right]+b_{o}\right) \\
		u &=\tanh \left(W_{u}\left[h_{i}^{l-1} ; x_{i} ; g^{l-1} ; \mathcal{N}_{i}^{l-1}\right]+b_{u}\right) \\
		c_{i}^{l} &=f_{i}^{l} \odot c_{i}^{l-1}+i_{i}^{l-1} \odot u \\
		h_{i}^{l} &=o_{i}^{l} \odot \tanh \left(c_{i}^{l}\right) \\
	\end{aligned}
\end{equation}
where $x_i$ in each layer is used to introduce the original meaning of the token; $\mathcal{N}_{i}$ represents the aggregated hidden vectors of the neighbors of the node $w_i$; $\left[ ; ; \right]$ represents the concatenation of the tokens' vectors; $i,f,o$ represent the input, forget and output gate structure respectively. 
% By adding the graph-level node $g$, the network is capable of modeling the interaction between each node, each edge, and the whole graph, which makes it better to focus on the more important information.

\subsubsection{Aggregation Module}
We assumes that $w_i$ has $C_i$ neighbors in the text graph. The aggregated hidden vectors $\mathcal{N}_{i}$ can be calculated as follows:
\begin{equation}\label{eq.3}
	\begin{aligned}
		\mathcal{P}_{i} &=h_{i}+p_{i} \\
		\alpha_{j} &=u\left(W_{n}\left[\mathcal{P}_{i} ; x_{i} ; g ; \mathcal{P}_{j}\right]+b_{n}\right) \\
		\mathcal{S}_{j} &=\frac{\exp \left(\alpha_{j}\right)}{\sum_{k}^{C_{i}} \exp \left(\alpha_{k}\right)} \\
		\mathcal{N}_{i} &=\sum_{k}^{C_{i}} \mathcal{S}_{k} h_{i_{k}}
	\end{aligned}
\end{equation}
where $p_{i}$ represents the positional vectors of the node $w_{i}$; $h_{i_{k}}$ represents the hidden state of the $k$-th neighbor of the node $w_{i}$. With positional vectors, it makes easier to aware of the position information of word $i$ and $j$ more seriously. The combination of the gated-graph neural network and the aggregation makes our model enable to gather the information from the long-term dependency as the layer number increases by determining which part of the information should be passed to higher layers.

The value of $g^{\ell}$ is updated according to the hidden states of the last layer $H^{\ell-1}$. $\widetilde{h}^{\ell-1}$ is calculated using attention mechanism as: $\left<h_1^{\ell},h_2^{\ell},\ldots,h_m^{\ell}\right>$.
\begin{equation}\label{eq.4}
	\begin{aligned}
		\alpha_{i}&=u\left(W_{a} h_{i}\right) \\
		\mathcal{S}_{i}&=\frac{\exp \left(\alpha_{i}\right)}{\sum_{j} \exp \left(\alpha_{j}\right)} \\
		\widetilde{h}&=\sum_{j} \mathcal{S}_{j} h_{j}
	\end{aligned}
\end{equation}
For each $h_i^{\ell-1}$, the forget gate $f_i^{\ell}$ is updated to decide which of the information should be considered to forget according to the vectors of the graph-node $g^{\ell}$. Also, the candidate vectors of the $c_g^{\ell}$ is updated according to the last-state candidate graph vectors $c_g^{\ell-1}$. The candidate vectors of the last-state node is denoted as $c_i^{\ell-1}$:
\begin{equation}\label{eq.5}
	\begin{aligned}
		\hat{f}_{g}^{l}&=\sigma\left(W_{g}\left[g^{l-1} ; \widetilde{h}^{l-1}\right]+b_{g}\right) \\
		\hat{f}_{i}^{l}&=\sigma\left(W_{f}\left[g^{l-1} ; h_{i}^{l-1}\right]+b_{f}\right) \\
		o^{l}&=\sigma\left(W_{o}\left[g^{l-1} ; \bar{h}^{l-1}\right]+b_{o}\right) \\
		f_{0}^{l}&, \cdots, f_{m}^{l}, f_{g}^{l}=\operatorname{Softmax}\left(\hat{f}_{0}^{l}, \cdots, \hat{f}_{m}^{l}, \hat{f}_{g}^{l}\right) \\
		c_{g}^{l}&=f_{g}^{l} \odot c_{g}^{l-1}+\sum_{i} f_{i}^{l} \odot c_{i}^{l-1} \\
		g^{l}&=o^{l} \odot \tanh \left(c_{g}^{l}\right)
	\end{aligned}
\end{equation}

\subsubsection{Trigger-Enhanced Decoding and Tagging}
Given the the vector of the whole graph $g^{L}$ and the vectors of each node $\mathbf{H}=\left\langle h_{1}^{L}, h_{2}^{L}, \cdots, h_{m}^{L}\right\rangle$, we use the previously trained module in \ref{sec:trigger-encoding and semantic-matching} to compute the mean value of $\hat{\mathbf{g}_{\mathbf{t}}}$ corresponds to its sentence. We incorporate the weighted sum of all the token representations from $\mathbf{H}$ as $\mathbf{H}^{'}$ with the trigger representation:
\begin{equation}
	\begin{aligned}
		\mathbf{H}^{'}&=\operatorname{SoftMax}\left(W^{T}\tanh \left(U_{1}\mathbf{H}^{T}+U_{2}\hat{\mathbf{g}_{\mathbf{t}}}^{T}\right)\right)^{T}\mathbf{H} \\
	\end{aligned}
\end{equation}
where $U_1, U_2$, and $W$ are trainable parameters. We concatenate $\mathbf{H}$ with the trigger-enhanced $\mathbf{H}^{'}$ as the input ($[\mathbf{H};\mathbf{H}^{'}]$) to the final CRF tagger. CRF tagger is constructed conventionally to predict the tag for each token according to~\cite{Liu2020}.
% Note that in this stage, our learning objective is the same as the conventional NER according to \cite{Lin2020}, which is to correctly predict the tag for each token.
It should be noted that When process the unlabeled sentences, we do not know the corresponding triggers of the sentence. Instead, we can use the trigger encoding and semantic matching module to compute the similarities between the sentence's representation and the trigger's representation according to L2-norm distances. The triggers with highest score will be used as the additional inputs to the last tagging process.

\section{Experiment Analysis}
\subsection{Datasets}
We first introduce the NER datasets used in the experiments. All the datasets are public and popular in NER task.
\begin{itemize}
	\item \textbf{CoNLL-2002 \& CoNLL-2003}: We use the English corpora from the CoNLL 2002 and 2003 shared task: language-independent named entity recognition \cite{TjongKimSang2000,TjongKimSang2003} with standard split.
	\item \textbf{JNLPBA}: We use a molecular biology dataset for identifying the technical terms \cite{Collier2004}. It contains various type of nested entities. Examples of such entities include the names of proteins, genes and their location of activity such as cells or organism. We use the same standard split according to \cite{Habibi2017}.
	\item \textbf{BC5CDR}: Another bio-medical domain dataset which is well-studied and popular in evaluating the performance of nested-NER \cite{Li2016}. We use the standard split.
\end{itemize}
Both JNLPBA and BC5CDR datasets involves a large number of nested entities in the bio-medical domain.

\subsection{Lexicon}
We use the lexicon generated on a corpus of article pairs from Gigaword \footnote{https://catalog.ldc.upenn.edu/LDC2003T05}, consisting around 4 millon articles. Use tensorflow as implementation \footnote{https://www.tensorflow.org/datasets/catalog/gigaword}. The embeddings of the lexicon words were pre-trained by Glove word embedding \cite{Pennington2014} and fine-tuned during training.

\subsection{Baselines}
Several state-of-the-art NER algorithms (sorted by the timeline) for effectiveness evaluation are listed below:
\begin{itemize}
	\item \textbf{Neural layered model}: It merged the output of the stacked LSTM layers to build new representation for detected entities. \cite{Ju2018}.
	\item \textbf{Lattice LSTM}: It raised a lattice-structure based LSTM model, which encodes a sequence of input characters as well as all the potential words matching a lexicon \cite{Zhang2018}.
	\item \textbf{Boundary-aware neural model}: It proposed a boundary-aware neural model for nested NER using sequence labelling \cite{Zheng2019}.
	\item \textbf{Biaffine-NER}: It proposed a graph-based dependency parsing to provide a global view on the sentence \cite{Yu2020}.
	\item \textbf{BiFlaG}: It firstly used the entities recognized by the flat NER module to construct an entity graph \cite{Luo2020}.
	\item \textbf{LGN}: It developed a lexicon-based graph network with global semantics and proposed a schema to connect the character using external lexicon words \cite{Gui2019}.
	\item \textbf{Trigger-NER}: It first introduced the concept of ``entity trigger'' and proposed a RNN-based trigger matching network for NER \cite{Lin2020}.
	\item \textbf{ACE $+$ document-context}: It proposed an automated concatenation method of embeddings (ACE) to automate the process of finding better concatenations of embeddings for structured prediction tasks, NER is one of them \cite{Wang2021}.
\end{itemize}

\subsection{Experiment Settings}
\subsubsection{Annotating Entity Triggers as Complementary Supervision}
We follow \cite{Lin2020} to crowd-source the entity triggers, and use the LEAN-LIFE developed by \cite{Lee2020} for annotating. Annotators are required to label a set of words in a sentence that are useful for entity recognition. We mask the entities with their corresponding types, so that human annotators can pay more attention to the non-entity words in the sentence. We merge multiple triggers for each entity, taking the intersection of all annotators' results. We reuse 14K triggers from \cite{Lin2020} and release another 8K triggers on the bio-domain for future trigger-enhanced NER in domain-research, which is also one of the contributions of our work.

\subsubsection{Hyper-parameter Settings}
We develop the Trigger-GNN based on Pytorch, with the same learning rate setting selected from $\{2e-5, 2e-4\}$ as \cite{Gui2019}. We use the Dropout \cite{Srivastava2014} with a rate of 0.4 for all th embedding layers and a rate of 0.3 for the aggregation model to reduce the overfitting. 
The embedding vector size and the hidden states were both set to 150. The initial word vectors is based on the Glove word embedding \cite{Pennington2014}. Steps of message passing $T$ are selected from $\{1,2,3,4,5,6\}$ which is detailed analyzed in \ref{exp.1}. We use the standard metrics for evaluation: Precision (P), Recall (R), and F1 score (F1).

\subsection{Evaluation and Discussion}
We demonstrate the performance of our Trigger-GNN on both general and bio domains, especially focus on the nested-NER. Results on the CoNLL-2002 \& 2003, JNLPBA and BC5CDR datasets are shown in Table \ref{tab:dataset1}, \ref{tab:dataset2}, and \ref{tab:dataset3}, respectively. Compared with the recent methods, our Trigger-GNN obtains the best results by a large margin. In particular, Trigger-GNN obtains 1.93\% and 3.52\% boosting than its baseline model LGN on the general domain dataset CoNLL-2002 \& 2003. 
And as shown in Table \ref{tab:dataset2}, and \ref{tab:dataset3}, Trigger-GNN obtains 3.41\% improvement on JNLPBA than LGN, and 5.32\% on BC5CDR. The observation from the main results on these four dataset demonstrates that our proposed model can adapt to the both general domain and bio domain, and perform better than the recent methods.

\begin{table}[htb]
	\centering
	\small
	\caption{Main results on CoNLL-2002 \& 2003}
	\label{tab:dataset1}
	\begin{tabular}{lcc}
		\toprule
		\textbf{Models}  & \textbf{CoNLL-2002} & \textbf{CoNLL-2003} \\
		\hline
		Neural layered model \cite{Ju2018} & 85.23\% & 85.13\%\\
		Lattic LSTM \cite{Zhang2018} & 90.34\%  & 91.28\% \\
		Boundary-aware model \cite{Zheng2019} & 87.58\% & 91.58\%\\
		Biaffine-NER \cite{Yu2020} & 91.38\% & 91.63\%\\
		BiFlaG \cite{Luo2020} & 92.54\%  & 92.67\% \\
		ACE+document-context \cite{Wang2021} & 93.95\% & 94.63\%\\
		\hline
		Trigger-NER \cite{Lin2020} & 86.83\% & 86.95\%\\
		LGN \cite{Gui2019} & 92.19\% & 91.86\% \\
		\textbf{Trigger-GNN} & \textbf{94.92\%} & \textbf{95.38\%}\\
		\hline
	\end{tabular}
\end{table}

\begin{table}[htb]
	\centering
	\small
	\caption{Main results on JNLPBA.}
	\label{tab:dataset2}
	\begin{tabular}{lccc}
		\hline
		\textbf{Models}  & \textbf{P} & \textbf{R} & \textbf{F1} \\
		\hline
		Neural layered model \cite{Ju2018} & 81.75\% & 81.24\% & 81.49\%\\
		Lattic LSTM \cite{Zhang2018} & 83.25\%  & 80.25\% & 81.72\% \\
		Boundary-aware model \cite{Zheng2019} & 80.12\% & 75.12\% & 77.53\%\\
		ACE+document-context \cite{Wang2021} & 85.75\% & 83.62\% & 84.67\%\\
		Trigger-NER \cite{Lin2020} & 79.12\% & 76.34\% & 77.70\%\\
		LGN \cite{Gui2019} & 84.12\% & 82.14\% & 83.12\%\\
		\textbf{Trigger-GNN} & \textbf{87.75\%} & \textbf{85.34\%} & \textbf{86.53\%}\\
		\hline
	\end{tabular}
\end{table}

\begin{table}[htb]
	\centering
	\small
	\caption{Main results on BC5CDR.}
	\label{tab:dataset3}
	\begin{tabular}{lccc}
		\hline
		\textbf{Models}  & \textbf{P} & \textbf{R} & \textbf{F1} \\
		\hline
		Neural layered model \cite{Ju2018} & 87.12\% & 86.14\% & 86.62\%\\
		Lattic LSTM \cite{Zhang2018} & 89.34\%  & 87.89\% & 88.61\% \\
		Boundary-aware model \cite{Zheng2019} & 86.12\% & 85.12\% & 85.62\%\\
		ACE+document-context \cite{Wang2021} & 93.65\% & 92.32\% & 92.98\%\\
		Trigger-NER \cite{Lin2020} & 85.75\% & 83.62\% & 84.67\%\\
		LGN \cite{Gui2019} & 90.12\% & 86.21\% & 88.13\%\\
		\textbf{Trigger-GNN} & \textbf{94.19\%} & \textbf{92.73\%} & \textbf{93.45\%}\\
		\hline
	\end{tabular}
\end{table}

\subsubsection{Steps of Message Passing}\label{exp.1}
\begin{figure}[htb]
	\centering
	\includegraphics[width=0.85\linewidth]{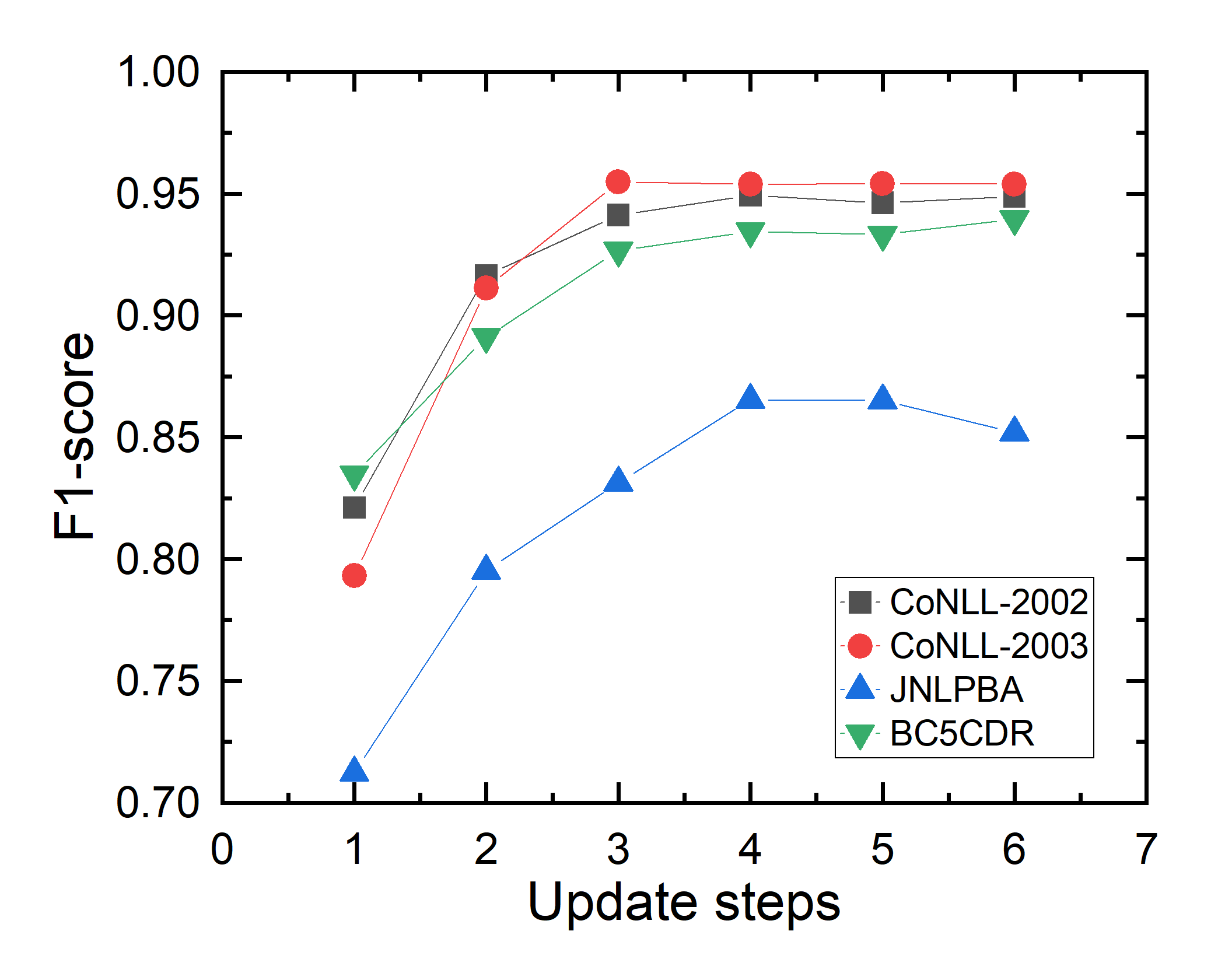}
	\caption{F1 variation under different update steps on the four datasets.}
	\label{fig:exp.2}
\end{figure}

To investigate the impact of steps of massage passing $T$ during the update process, we analyze the performance of Trigger-GNN using different value of the step number as shown in Fig.~\ref{fig:exp.2}. 
The results show that the number of update steps has an important impact on the performance of Trigger-GNN. When $T$ is less than 3, the F1 score drops by 4.1\% on average. Specifically,  F1 score on either JNLPBA and BC5CDR datasets drops around 3.63\% and 3.53\%. 
Several rounds of updates to the model yield competitive results, revealing that the Trigger-GNN has benefitted from the update process.
Empirically, as the process iterates, graph nodes aggregate more information from the neighbors and graph-level node as it aggregates information from both the neighbors' nodes and the edges at every update step.
% In the Trigger-GNN model, more valuable information can be captured through the recursive aggregation.

\subsubsection{Ablation Studies}\label{exp.2}
To evaluate the contribution of each component in Trigger-GNN, we conduct the ablation study on four public datasets. The results are as illustrated in Table \ref{tab:ablation study}.
It shows that the model's performance drops when the graph-level node is removed, which indicates the essential of the global connections in the graph structure. We can also observe that the entity triggers play an vital role. Specially, the CoNLL-2003, BC5CDR, and JNLPBA suffer serious performance drops of 3\% without entity triggers. Also, missing the edge/lexicon will result in a further performance loss about 1.5\% on average.

\begin{table}[htb]
	\centering
	\small
	\caption{An ablation study of Trigger-GNN. F1 scores were evaluated.}
	\label{tab:ablation study}
		\begin{tabular}{lcccc}
			\hline
			\textbf{Models} & \textbf{CoNLL-2003} & \textbf{JNLPBA} & \textbf{BC5CDR} \\
			\hline
			Trigger-GNN  & 95.38\% & 86.53\% & 93.45\% \\
			\hspace{1em}-graph-level node   & 94.31\% & 85.87\% & 92.34\% \\
			\hspace{1em}\textbf{-trigger}  & \textbf{91.86\%} & \textbf{83.12\%} & \textbf{88.13\%} \\
			\hspace{1em}-edge/lexicon  & 90.42\% & 81.74\% & 87.83\% \\
			\hspace{1em}-bidirectional  & 88.12\% & 76.51\% & 84.38\%\\
			\hspace{1em}-crf  & 86.87\% & 74.93\% & 82.80\%\\
			\hline
			LGN  & 91.86\% & 83.12\% & 88.13\% \\
			\hspace{1em}-graph-level node & 89.73\% & 81.56\% & 87.32\% \\
			\hline
	\end{tabular}
\end{table}

To better demonstrate the advance of our model, we compare our trigger-based GNN with LGN using lexicon instead. The results show that compared to the LGN, the Trigger-GNN achieves an average F1 score of 2.68\% higher. In addition, there is a distinct performance gap when remove the global node from both of the models. This is because the trigger-based GNN can add in complementary supervision on the words, entity triggers and the whole sentence. The F1 score of LGN decreases by 1.83\% on average across the four datasets without graph-level nodes. In contrast, our Trigger-GNN drops by only 0.89\% which proves that Trigger-GNN is better at modeling sentences.

\subsubsection{Performance Against Labeled Data}\label{exp.3}
\begin{figure}[htb]
	\centering
	\includegraphics[width=\linewidth]{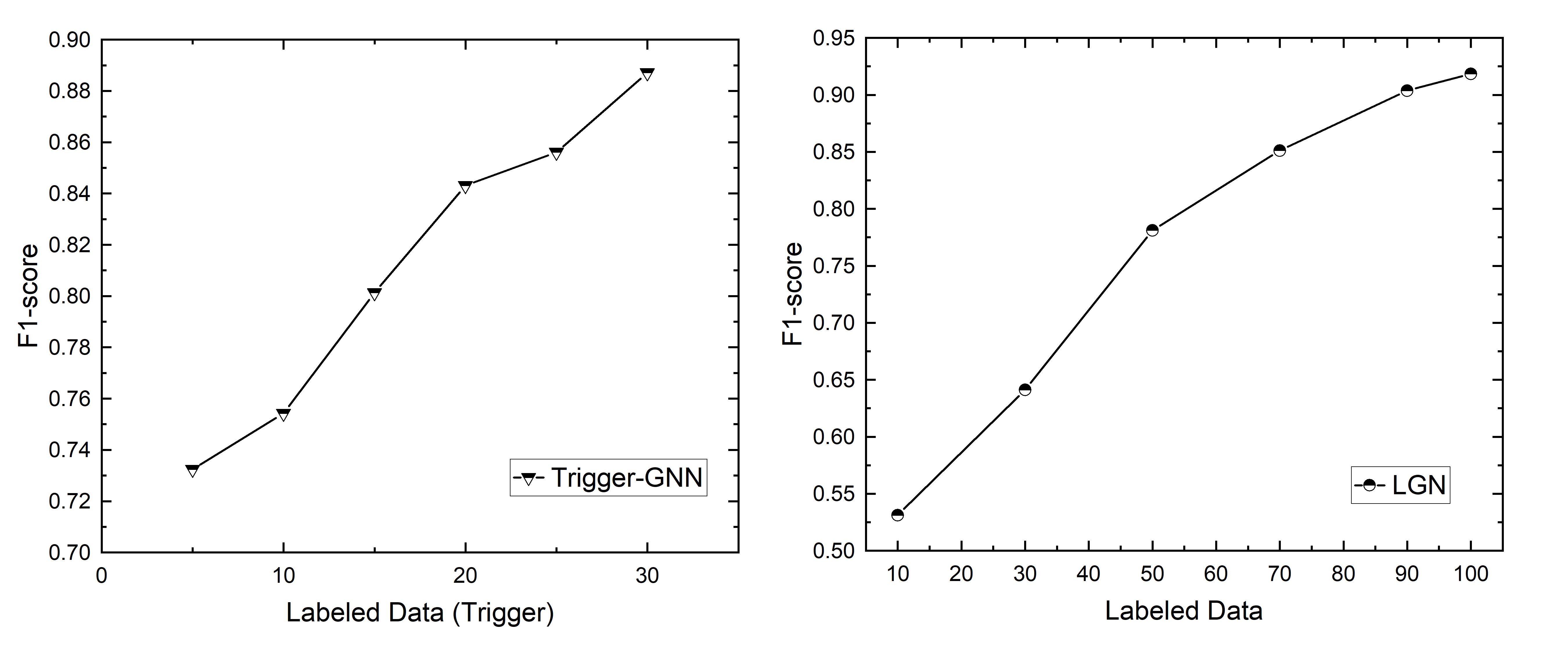}
	\caption{F1 score against labeled data on CoNLL-2003.}
	\label{fig:exp.3}
\end{figure}

Fig.~\ref{fig:exp.3} illustrates the performance of Trigger-GNN and several baseline methods on the CoNLL2003 dataset using different number of the labeled data. We can see that by using only 20-30\% of the trigger-annotated data for training, Trigger-GNN model delivers comparable performance as the baseline model LGN using 50-70\% traditional training data like lexicon. It indicates the cost-effectiveness of using triggers as an additional source of supervision.

\subsubsection{Case Studies}\label{exp.4}
To further validate the Trigger-GNN can alleviate the nested NER, we perform an case study on BC5CDR as illustrated in Table~\ref{tab:case study}. It demonstrates that the Trigger-GNN performs well on the nested NER case. In particular, Trigger-GNN can not only identify the ``selegiline'' as a chemical, but also can integrate the ``supine systolic and diastolic blood pressures'' as a integral disease. However, models like LGN and Trigger-NER can only capture part of the entities or incorrectly split the diseases, which illustrates the superiority of our proposed model.

\begin{table}[htb]
	\centering
	\small
	\caption{Case studies on BC5CDR.}
	\label{tab:case study}
	\begin{tabular}{ll}
		\hline
		\textbf{Models} & \multicolumn{1}{c}{\textbf{Cases}} \\ \hline
		LGN             & \begin{tabular}[c]{@{}l@{}}Stopping selegiline also significantly reduced the \\ {\color{blue}supine systolic and diastolic blood pressures} \\ {\color{blue}(Disease)} consistent with a previously \\supine pressure action.\end{tabular}  \\
		Trigger-NER     & \begin{tabular}[c]{@{}l@{}}Stopping {\color{red}selegiline (Chemical)} also significantly\\ reduced the {\color{blue}supine systolic (Disease)} and diastolic \\ blood pressures consistent with a previously \\ supine pressure action.\end{tabular} \\
		Trigger-GNN     & \begin{tabular}[c]{@{}l@{}}Stopping {\color{red}selegiline (Chemical)} also significantly\\ reduced the {\color{blue}supine systolic and diastolic blood} \\ {\color{blue} pressures (Disease)} consistent with a previously\\ supine pressure action.\end{tabular}  \\
		\hline
	\end{tabular}
\end{table}

\section{Conclusion}
In this work, we investigate a trigger-based graph neural network approach to alleviate the nested NER. Entity triggers are used to provide more explicit supervision. The Trigger-GNN enables two significant issues: (1) multiple graph-based interactions among the words, entity triggers, and the whole sentence semantics; and (2) using trigger-based instead of lexicon-based \cite{Gui2019} can add in complementary supervision signals and thus help the model to learn and generalize more efficiently.
As a result, the experiments indicate the significant performance of our proposed model on four real-world datasets in both general and bio domains. The explanatory experiments also illustrate the efficiency and the cost-effectiveness of our proposed model.

\section{Acknowledgment}
This work is supported by the National Natural Science Foundation of China (62002207, 62072290, 12075142, 62073201), the Shandong Provincial Natural Science Foundation (ZR2020MA102) and Shandong Provincial Key Laboratory for Novel Distributed Computer Software Technology.

\bibliographystyle{IEEEtran}
\bibliography{IEEEabrv,ref}

\end{document}